\documentclass[letterpaper, 10 pt, conference]{ieeeconf}

\IEEEoverridecommandlockouts

\usepackage[T1]{fontenc}
\usepackage[utf8]{inputenc}
\usepackage[english]{babel}
\usepackage{textcomp}
\usepackage{url}
\usepackage{booktabs, tabularx}
\usepackage{amsfonts}
\usepackage{nicefrac}
\usepackage{times}
\usepackage{siunitx}
\usepackage[plain]{fancyref}
\usepackage{calc}
\usepackage{etoolbox}
\usepackage{lipsum}

\usepackage{balance}
\usepackage{hyperref}
\usepackage{url}

\usepackage{amsthm}
\usepackage{amssymb}
\usepackage{mathtools}
\usepackage{mathrsfs}

\usepackage{graphics}

\usepackage{tikz, pgfplots, pgfplotstable}

\usepackage{pgfkeys}
\usepackage{standalone}
\usepackage{cite}
\usepackage{graphicx}
\usepackage{multirow}
\usepackage{threeparttable}
\usepackage{placeins}
\usepackage{adjustbox}
\usepackage[super]{nth}
\usepackage[size=footnotesize]{caption}
\usepackage[size=footnotesize]{subcaption}
\usepackage{flushend}
\usepackage[symbol]{footmisc}
\usepackage{nopageno}

\usepackage{dsfont}
\usepackage{scalerel}

\makeatletter
\let\NAT@parse\undefined
\makeatother

\usepackage[numbers,sort&compress]{natbib}
\bibpunct{[}{]}{,}{n}{}{;}

\DeclareSIUnit\px{px}

\tikzset{>=latex}

\pgfdeclarelayer{background}
\pgfsetlayers{background,main}

\makeatletter
\newcommand\currentcoordinate{\the\tikz@lastxsaved,\the\tikz@lastysaved}
\makeatother

\pgfplotsset{compat=1.14}
\usetikzlibrary{shapes, shapes.geometric, arrows, calc, 3d, chains, arrows.meta, bending, quotes}
\usepgfplotslibrary{colorbrewer, colormaps}

\pgfplotsset{every axis/.style = {cycle list/Spectral}}

\newcommand{\quotes}[1]{``#1''}

\NewDocumentCommand\optimal{m e_}{
  \IfNoValueTF{#2}{
    {#1}^\ast
  }{
    {#1}^\ast_{#2}
  }
}

\definecolor{mygreen}{HTML}{D2691E}
\definecolor{myred}{HTML}{ee0000}
\definecolor{mysepia}{HTML}{704214}
\definecolor{turquoise}{HTML}{10d59c}

\usepackage{pifont}

\usepackage{acronym}
\acrodef{DTA}{Duckietown Autolab}
\acrodef{AI-DO}{AI Driving Olympics}
\acrodef{LF}{lane-following}
\acrodef{DUCKIENet}{Decentralized Urban Collaborative Benchmarking Network}
\acrodef{MPD}{Mean Position Deviation}
\acrodef{MOD}{Mean Orientation Deviation}

\usepackage{ifxetex}
\ifxetex

\usepackage[protrusion=true,tracking=false,kerning=false,spacing=false]{microtype}
\else

\usepackage[tracking=false,kerning=true,spacing=true]{microtype}

\fi
\renewcommand{\baselinestretch}{0.98}
\let\labelindent\relax
\usepackage{enumitem}
\setlist[enumerate]{itemsep=0mm,leftmargin=4mm,topsep=0mm}
\setlist[itemize]{itemsep=0mm,leftmargin=4mm,topsep=0mm}

\usetikzlibrary{external}
\newcommand{\expfigure}[3]{%
{\centering%
    \begin{minipage}[c]{0.5\columnwidth}
        \centering
        \includegraphics[height=4.6cm,width=\textwidth,keepaspectratio=true]{#1}
    \end{minipage}%
    \hfill
    \begin{minipage}[c]{0.5\columnwidth}
        \centering
        \includegraphics[height=5cm,width=\textwidth,keepaspectratio=true]{#2}
    \end{minipage}%
    \vspace{1mm}

    \footnotesize
    \begin{tabular}{@{}cccccccc@{}}
    \toprule
 Robots & Location & Repetitions & \begin{tabular}[c]{@{}c@{}}AVG\\ MPD\end{tabular} & \begin{tabular}[c]{@{}c@{}}STD\\ MPD\end{tabular} & \begin{tabular}[c]{@{}c@{}}AVG\\ MOD\end{tabular} & \begin{tabular}[c]{@{}c@{}}STD\\ MOD\end{tabular} \\ \midrule
    #3 \bottomrule
    \end{tabular}

}
}

\newcommand{\exptablesamebot}{
 1 & ETHZ & 9 & -6.2 & 1.3 & -0.8 & 2.8 \\
}

\newcommand{\exptablediffbot}{
 3 & ETHZ & $3\times3$ & 1.1 & 3.4 & -0.4 & 5.2 \\
}

\newcommand{\exptablediffsub}{
 $1 \times 2$ & ETHZ/TTIC & $6\times2$ & 2.2 & 2.5 & -0.4 & 3.9 \\

}

\title{\LARGE \bf Integrated Benchmarking and Design for Reproducible\\ and Accessible Evaluation of Robotic Agents}

\author{ \small
Jacopo Tani,${}^{1,*}$
Andrea F.\ Daniele,${}^{2}$
Gianmarco Bernasconi,${}^1$
Amaury Camus,${}^1$
Aleksandar Petrov,${}^1$
Anthony Courchesne,${}^{3}$ \\
Bhairav Mehta,${}^{3}$
Rohit Suri,${}^1$
Tomasz Zaluska,${}^1$
Matthew R.\ Walter,${}^2$
Emilio Frazzoli,${}^{1,4}$
Liam Paull,${}^{3}$
Andrea Censi${}^{1}$
  \thanks{${}^1$\,ETH Z\"urich, Switzerland;
${}^2$\,Toyota Technological Institute at Chicago, Chicago, USA;
${}^3$\,Mila, Universit\'e de Montr\'eal, Montr\'eal, Canada;
\mbox{${}^4$\,nuTonomy}, an Aptiv company, Boston, USA;
\mbox{${}^*$\,Corresponding author: tanij@ethz.ch.}
         }
 }

\begin{document}
\maketitle
\thispagestyle{plain}
\pagestyle{plain}

\begin{abstract}
As robotics matures and increases in complexity, it is more necessary than ever that robot autonomy research be \textit{reproducible}. Compared to other sciences, there are specific challenges to benchmarking autonomy, such as the complexity of the software stacks, the variability of the hardware and the reliance on data-driven techniques, amongst others. In this paper, we describe a new concept for reproducible robotics research that integrates development and benchmarking, so that reproducibility is obtained \quotes{by design} from the beginning of the research/development processes. We first provide the overall conceptual objectives to achieve this goal and then a concrete instance that we have built: the DUCKIENet. One of the central components of this setup is the Duckietown Autolab, a remotely accessible standardized setup that is itself also relatively low-cost and reproducible.
When evaluating agents, careful definition of interfaces allows users to choose among local versus remote evaluation using simulation, logs, or remote automated hardware setups.
We validate the system by analyzing the repeatability of experiments conducted using the infrastructure and show that there is low variance across different robot hardware and across different remote labs.\footnote[2]{The code used to build the system is available on the Duckietown GitHub page (\url{https://github.com/duckietown}).}
\end{abstract}

\section{Introduction}
\label{sec:introduction}

Mobile robotics poses unique challenges that have precluded the establishment of benchmarks for rigorous evaluation. Robotic systems are \emph{complex} with many different interacting components. As a result, evaluating the individual components is not a good proxy for full system evaluation. Moreover, the outputs of robotic systems are temporally correlated (due to the system dynamics) and partially observable (due to the requirement of a perception system). Disentangling these issues for proper evaluation is daunting.

\begin{figure}[tb]
    \centering
    \includegraphics[width=0.95\columnwidth, trim = 0 0 0 0, clip]{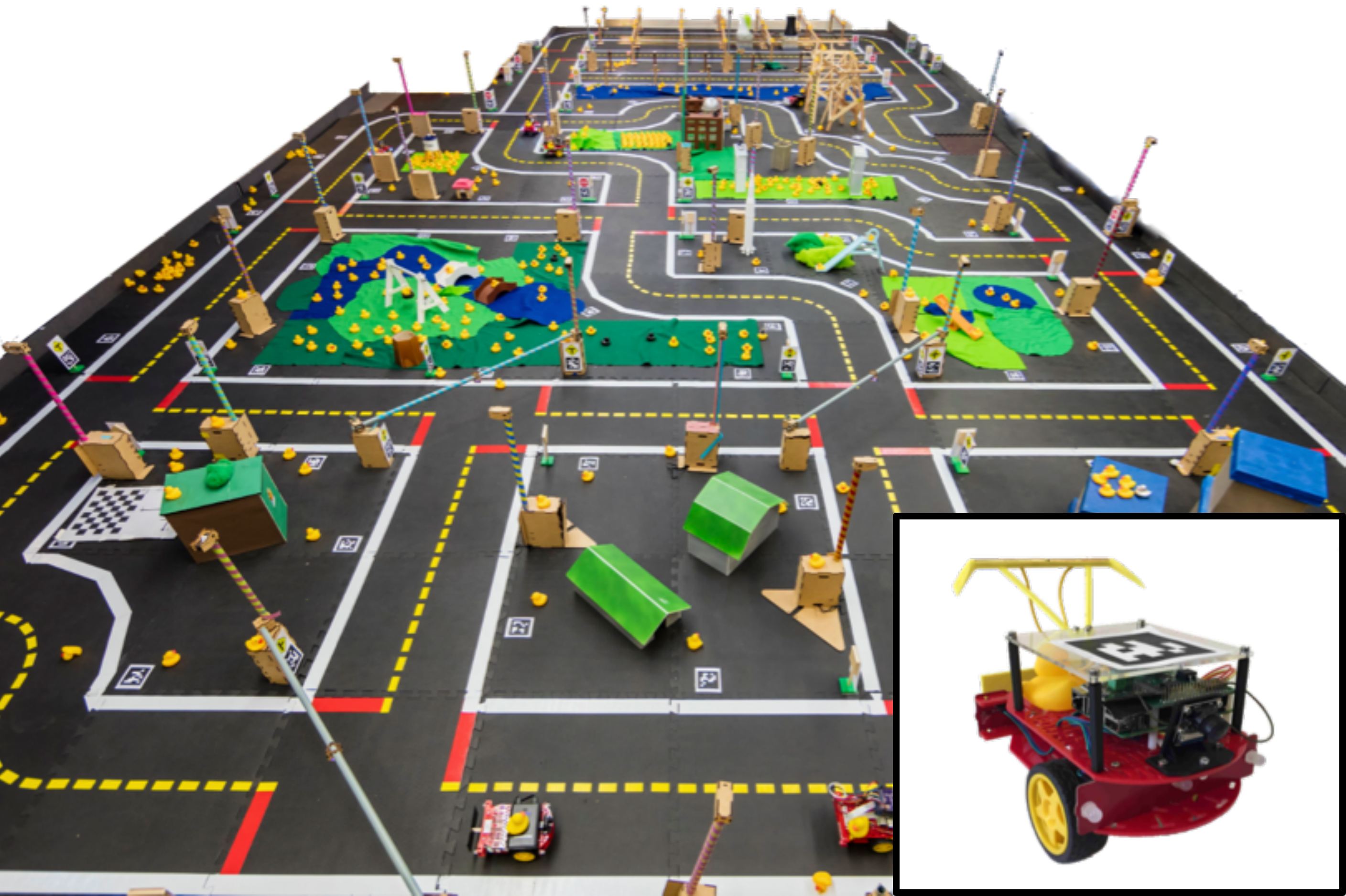}
    \caption{\textbf{The Duckietown Autolab}: We augment Duckietowns~\cite{paull2017duckietown} with localization and automatic charging infrastructure. We modify Duckiebots (inset) to facilitate detection by the localization system and auto-charging.}\label{fig:db18-autolab}
\end{figure}
The majority of research on mobile robotics takes place in idealized laboratory settings or in unique uncontrolled environments that make comparison difficult. Hence, the value of a specific result is either open to interpretation or conditioned on specifics of the setup that are not necessarily reported as part of the presentation.
The issue of reproducibility is exacerbated by the recent emergence of \emph{data-driven} approaches, the performance of which can vary dramatically and unpredictably across seemingly identical environments (e.g., by only varying the random seed~\cite{Henderson2017}). It is increasingly important to fairly compare these data-driven approaches with the more classical methods or hybrids of the two.

Most existing methods for evaluating robotics operate on individual, isolated components of the system. For example, evaluating robot perception is comparatively straightforward and typically relies on annotated datasets~\cite{kitti, cityscapes, nuscenes, waymo-sun2019scalability}. However, performance on these benchmark datasets is often not indicative of an algorithm's performance in practice.
A common approach to analyzing robot control algorithms is to abstract away the effects of perception~\cite{yu2020introduction, robotarium} and assume that the pose of the robot is known (e.g., as determined using an external localization system, such as a motion capture setup~\cite{vicon-merriaux2017study}).

Simulation environments are potentially valuable tools for system-level evaluation. Examples such as CARLA~\cite{dosovitskiy2017carla}, AirSim~\cite{shah2018airsim}, and Air Learning~\cite{krishnan2019air} have recently been developed for this purpose. However, a challenge with simulation-based evaluation is that it is difficult to quantify the extent to which the results extend to the real world.

Robotics competitions have been excellent testbeds for more rigorous evaluation of robotics algorithms~\cite{amigoni2015competitions}. For example, the DARPA urban challenge~\cite{buehler2009darpa}, the DARPA robotics challenge~\cite{johnson2015team}, and the Amazon Picking Challenge~\cite{correll2016analysis}, have all resulted in massive development in their respective sub-fields (autonomous driving, humanoid robotics, and manipulation respectively). However, the cost and multi-year effort required to join these challenges limit participation.

A very promising recent trend that was spearheaded by the Robotarium at Georgia Tech~\cite{pickem2017robotarium, wilson2020robotarium} is to provide remotely accessible lab setups for evaluation. This approach has the key advantage that it enables access for anyone to submit. In the case of the Robotarium, the facility itself cost 2.5M \$US, and would therefore be difficult to replicate. Additionally, while it does allow users flexibility in the algorithms they can run, it does not offer any standardized evaluation.

 In this paper, we propose to harmonize all the above-mentioned elements (problem definition, benchmarks, development, simulation, annotated logs, experiments, etc.) in a \quotes{closed-loop} design that has minimal software and hardware requirements. This framework allows users to define benchmarks that can be evaluated with different modalities (in simulation or on real robots) locally or remotely. Importantly, the design provides immediate feedback from the evaluations to the user in the form of logged data and scores based on the metrics defined in the benchmark.

We present the \ac{DUCKIENet}, an instantiation of this design based on the Duckietown platform~\cite{paull2017duckietown} that provides an accessible and reproducible framework focused on autonomous vehicle fleets operating in model urban environments.
The \ac{DUCKIENet} enables users to develop and test a wide variety of different algorithms using available resources (simulator, logs, cloud evaluations, etc.), and then deploy their algorithms locally in simulation, locally on a robot, in a cloud-based simulation, or on a real robot in a remote lab. In each case, the submitter receives feedback and scores based on well-defined metrics.

The \ac{DUCKIENet} includes \acfp{DTA}, remote labs that are also low-cost, standardized and fully documented with regards to assembly and operation, making this approach highly scalable. Features of the \acp{DTA} include an off-the-shelf camera-based localization system and a custom automatic robot recharging infrastructure. The accessibility of the hardware testing environment through the \ac{DUCKIENet} enables experimental benchmarking that can be performed on a network of \acp{DTA} in different geographical locations.

In the remainder of this paper we will summarize the objectives of our integrated benchmarking system in Section~\ref{sec:preliminaries}, describe an instantiation that adheres to these objectives, the DUCKIENet, in Section~\ref{sec:dn}, validate the performance and usefulness of the approach in Section~\ref{sec:dn-validation} and finally present conclusions in Section~\ref{sec:conclusion}.

\section{Integrated benchmarking and development for reproducible research}
\label{sec:preliminaries}

\begin{figure}[htb]
\centering
    \includegraphics[width=0.95\columnwidth]{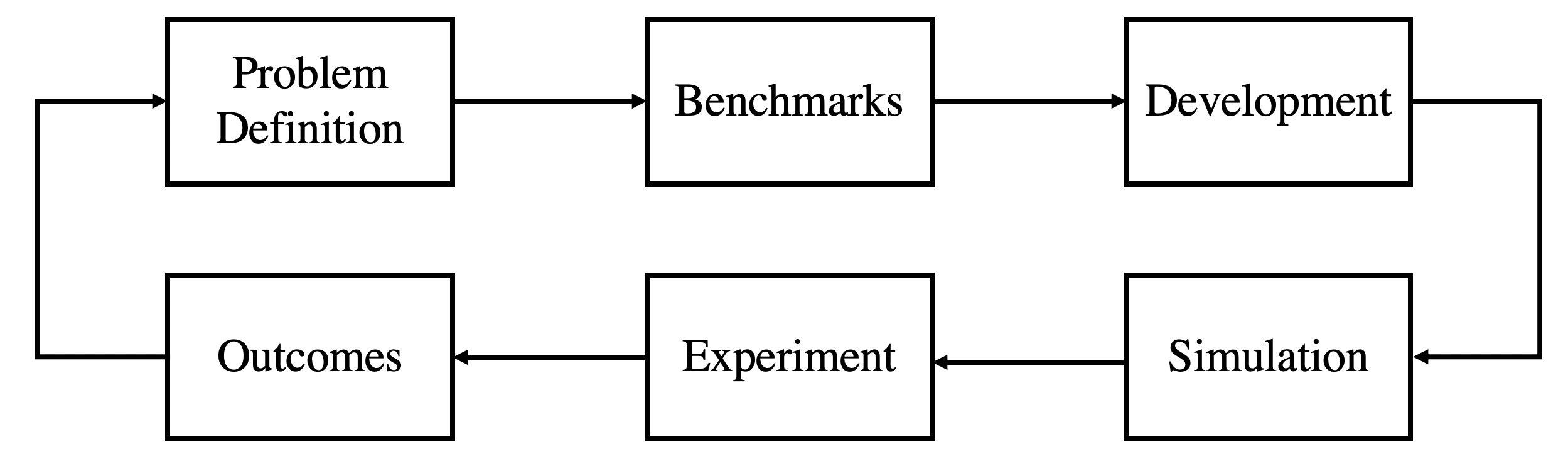}
     \caption{\textbf{Integrating Benchmark Design}: Designing robotic benchmarks should be more tightly integrating into robotic system development.}
    \label{fig:int-loop}
\end{figure}

Benchmarking should not be considered an isolated activity that occurs as an afterthought of development. Rather, it should be considered as an integral part of the research and development process itself, analogous to test-driven development in software design.
Robotics development should be a feedback loop that includes problem definition, the specification of benchmarks, development, testing in simulation and on real robot hardware, and using the results to adjust the problem definition and the benchmarks (Fig.~\ref{fig:int-loop}).

In this paper we describe a system built to explore the benefits and the challenges of harmonizing development, benchmarking, simulation and experiments to obtain reproducibility \quotes{by design}.
In this section we give an overview of the main challenges and design choices.

\subsection{Reproducibility}

Our main objective is to achieve \quotes{reproducibility}. At an abstract level, reproducibility is the ability to reduce variability in the evaluation of a system such that experimental results are more credible~\cite{Goodman341ps12}.
More specifically, there are three distinct aspects to reproducibility, all equally important:
\begin{enumerate}
    \item An \emph{experiment} is reproducible if the results obtained at different times and by different people are similar;
    \item An \emph{experimental setup} is reproducible if it is possible to build a copy of the same setup elsewhere and obtain comparable results;
    \item  An \emph{experimental platform} is reproducible if it is relatively easy, in terms of cost and complexity, to replicate.
\end{enumerate}

\subsubsection{Software}

Modern autonomy stacks consist of many modules and dependencies, and the reproducibility target implies being able to use the same software possibly many years later.

Standardized software archival systems such as Git solve part of the problem, as it is easy to guarantee storage and integrity of the software over many years~\cite{blischak2016quick}. However, even if the core implementation of the algorithm is preserved, the properties of the system change significantly in practice due to the external dependencies (libraries, operating system, etc.). In extreme cases \quotes{bit rot} causes software to stop working simply because the dependencies are no longer maintained~\cite{boettiger2015introduction}.

Containerization technologies such as Docker~\cite{Docker} have dramatically reduced the effort required to create reproducible software. Containerization solves the problem of compilation reproducibility, as it standardizes the environment in which the code runs~\cite{weisz2016robobench}.
Furthermore, the ability to easily store and retrieve particular builds (\quotes{images}) through the use of registries, solves the problem of effectively freezing dependencies.  This allows perfect reproducibility for a time span of 5--10 years (until the images can no longer be easily run using the current version of the containerization platform).

\subsubsection{Hardware}
In addition to issues related to software, a central challenge in the verification and validation of robot autonomy is that the results typically depend on under-defined properties of the particular hardware platforms used to run the experiments. While storage and containerization make it possible to standardize the version and build of software, hardware will never be fully reproducible, and using only one instance of a platform is impossible due to wear and tear.

As a result, we propose to make several samples from a \emph{distribution} of hardware platforms that are used for evaluation. Consequently, metrics should be considered statistically  (e.g., via mean and covariance), similarly to recent standard practices in reinforcement learning that evaluate over a set of random seeds~\cite{Henderson2017}. This incentivizes algorithms that are robust to hardware variation, and will consequently be more likely to work in practice. Also, one can define the distribution of variations in simulations such that configurations are sampled by the benchmark on each episode.
While the distribution as a whole might change (due to different components being replaced over the years), the distributional shift is easier to detect by simply comparing the performance of an agent on the previous and current distributions.

Cost can also quickly become the limiting factor for the reproducibility of many platforms. Our objective is to provide a robotic platform, a standardized environment in which it should operate, and a method of rigorous evaluation that is relatively inexpensive to acquire and construct.

\subsubsection{Environment} We use an analogous approach to deal with variability in the environment. The first and most crucial step for reproducibility of the setup is to standardize the formal description of the environment. Nevertheless, it is infeasible to expect that even a well described environment can be exactly reproduced in every instance.
As discussed in Section~\ref{sec:dn}, our approach is to use a distributed set of experimental setups (the \acp{DTA}) and test agents across different installations that exist around the world. Similar to the way that robot hardware should be randomized, proper evaluation should include sampling from a distribution over environments.

\subsection{Agent Interoperability}
\label{sec:interoperability}

Open source robotics middleware such as ROS~\cite{ROS} and LCM~\cite{huang10} are extremely useful due to the tools that they provide. However, heavily leveraging these software infrastructures is an impediment to long-term software reproducibility since versions change often and backward/forward compatibility is not always ensured.
As a result, we propose a framework based on low-level primitives that are extremely unlikely to become obsolete, and instead provide \emph{bridges} to popular frameworks and middlewares. Specifically, components in our infrastructure communicate using pipes (FIFOs) and a protocol based on Concise Binary Object Representation (CBOR) (a binarized version of JavaScript Object Notation)~\cite{cbor}. These standardized protocols  have existed  for decades and will likely persist long into the future.
It is then the job of the infrastructure to interconnect components, either directly or using current middleware (e.g., ROS) for transport over networks. In the future, the choice of middleware might change without invalidating the stored agents as a new bridge can be built to the new middleware. An additional benefit of CBOR is that it is self-describing (i.e., no need of a previously agreed schema to decode the data) and allows some form of extensibility since schema can change or be updated and maintain backwards compatibility.

\subsection{Robot-Agent Abstraction}

Essential to our approach is defining the interface between the \emph{agent} and the \emph{robot} (whether the robot is in a simulator or real hardware). Fundamentally, we define this interface in the following way (Fig.~\ref{fig:interfaces}):
\begin{itemize}
\item \textbf{Robots} are nodes that consume actuation commands and produce sensor observations.
\item \textbf{Agents} are nodes that consume sensor observations and produce actuation commands.
\end{itemize}

Combining this with containerization and the low-level reproducible communication protocol introduced in \mbox{Section~\ref{sec:interoperability}} results in a framework that is reliable and enables  the easy exchange of either  agents or robots in a flexible and straightforward manner. As described in Section~\ref{sec:dn}, this approach is leveraged in the DUCKIENet to enable zero friction transitions from local development, to remote simulation, and to remote experiments in \acp{DTA}.

\subsection{Robot-Benchmark Abstraction}

\begin{figure}[tb]
\centering
    \includegraphics[trim = 0 0 0 -20, width=0.95\columnwidth]{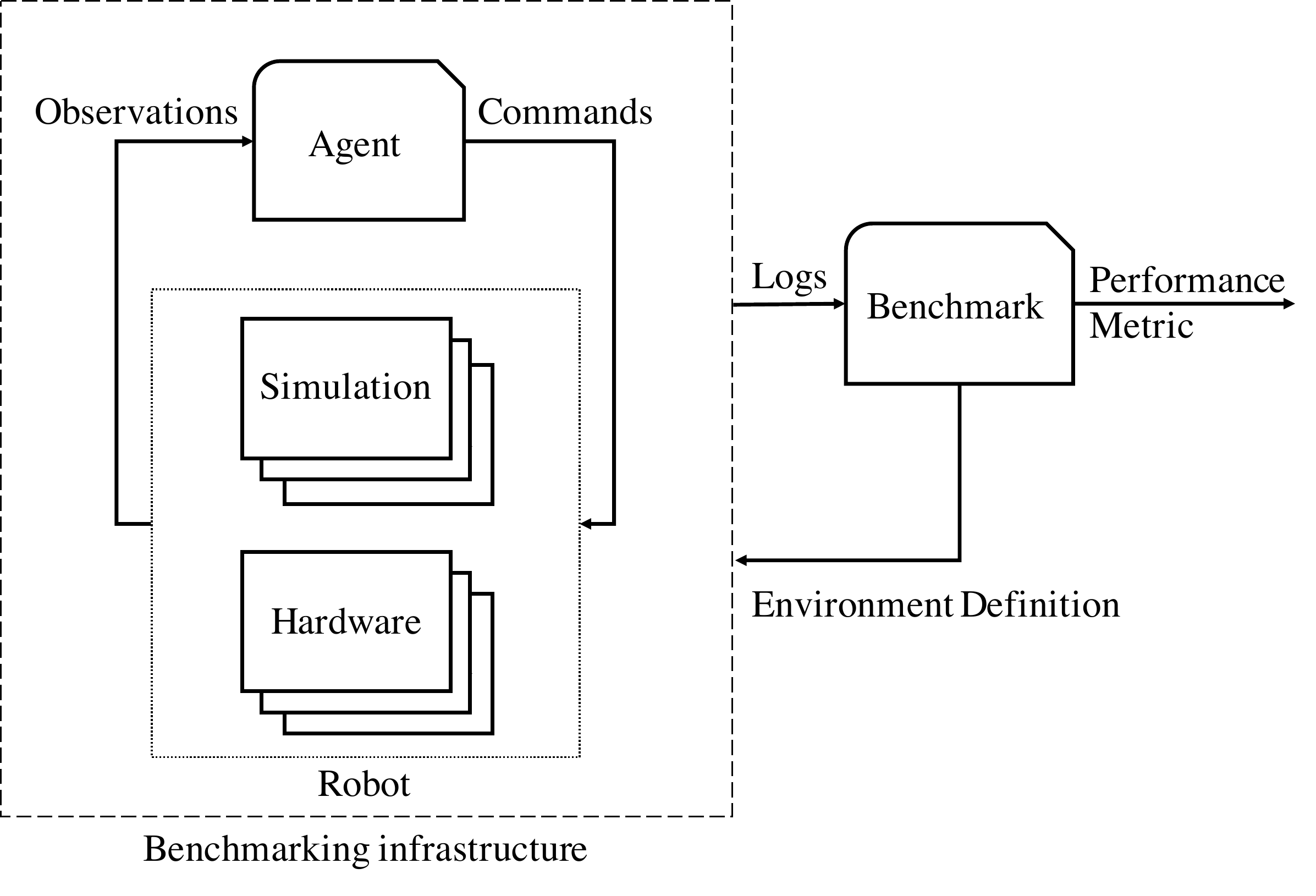}
     \caption{\textbf{Abstractions and Interfaces}: An interface must be well-specified so that components on either side of the interface can be swapped in and out. The most important is the Agent-Robot interface, where the robot can be real hardware or a simulator. The interface between the benchmark and the infrastructure enables a clear definition of benchmark tasks (in terms of metrics) separately from the means by which they are evaluated.}
    \label{fig:interfaces}
\end{figure}

A benchmarking system becomes very powerful and extensible if there is a way to distinguish the infrastructure from the particular benchmark being run. In similar fashion to the definition of the interface for the robot system, we standardize the definition of a benchmark and its implementation as containers (Fig.~\ref{fig:interfaces}):
\begin{itemize}
    \item \textbf{Benchmarks} provide environment configurations and consume robot states to evaluate well-define metrics.
    \item \textbf{Infrastructure} consumes environment configurations and provides estimates of robot state, either directly (e.g., through a simulator) or through other external means.
\end{itemize}

\section{The \ac{DUCKIENet}}
\label{sec:dn}

In this section we describe an instantiation of the high-level objectives proposed in Section~\ref{sec:preliminaries} that we refer to as \ac{DUCKIENet}, which:
\begin{itemize}
    \item comprises affordable robot hardware and environments,
    \item adheres to the proposed abstractions to allow easy evaluation on local hardware or in simulation,
    \item enables evaluation of user-specified benchmarks in the cloud or in a remote, standardized lab.
\end{itemize}

\begin{figure}[tb]
\centering
    \includegraphics[width=\columnwidth]{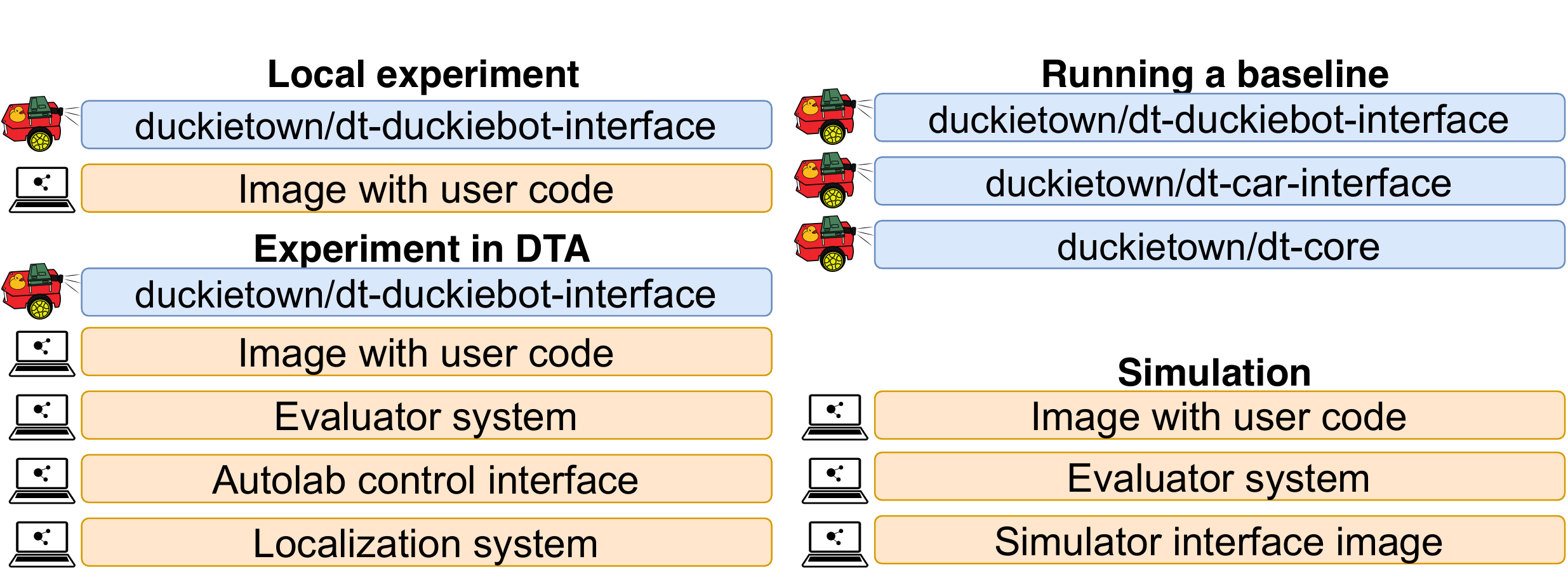}
    \caption{\textbf{Use-cases}: The well-defined interfaces enable execution of an agent in different settings, including locally in hardware, in a \ac{DTA}, with a standard baseline, or in a simulator. Different icons indicate where a container is executed (either on the robot
    \protect\includegraphics[height=0.85em]{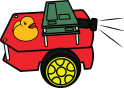} or on the laptop \protect\includegraphics[height=0.85em]{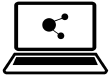}).}
    \label{fig:dt-usecases}
\end{figure}

\begin{figure}[tb]
\centering
    \includegraphics[width=\columnwidth]{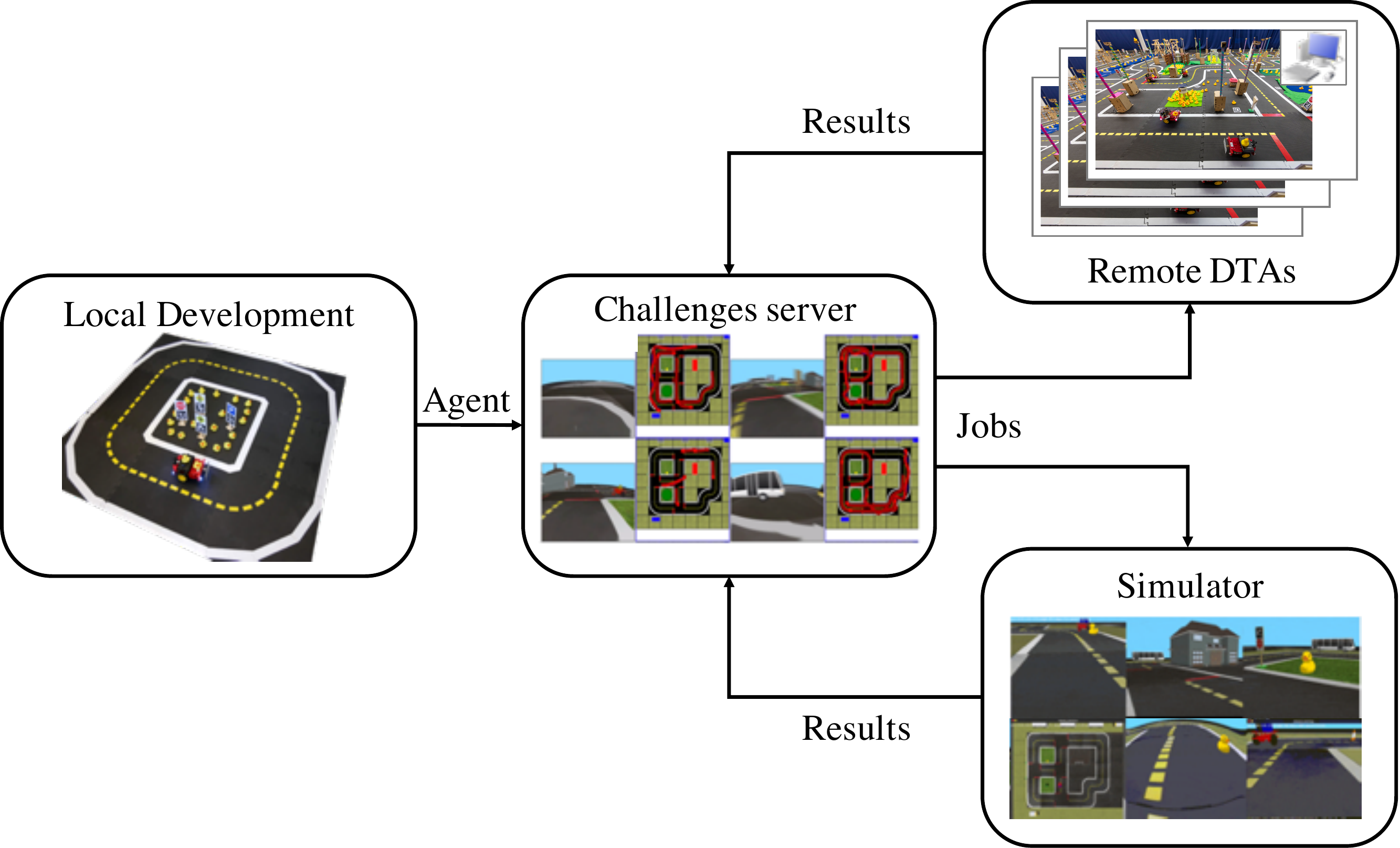}
    \caption{\textbf{The user workflow}: Local development can take place on the real hardware or in a simulator. Agents are submitted to the Challenges server, which marshals agents to be submitted in simulation challenges to the cloud simulation, or agents that should be evaluated on real hardware to a \ac{DTA}. The results of these evaluations are combined with the benchmark definition and made available to the submitter.}
    \label{fig:duckienet-high-level}
\end{figure}

\subsection{The Base Platform}

The \ac{DUCKIENet} is built using the Duckietown platform, which was originally developed in 2016 and has since been used for education~\cite{tani2016duckietown}, research~\cite{aido} and outreach~\cite{dt-org}.

\subsubsection{The Duckietown hardware platform} \label{sec:dt-hw}

The Duckietown hardware consists of Duckiebots and Duckietown urban environments. Duckiebots are differential-drive mobile robots with a front-facing camera as the only sensor, five LEDs used for communication, two DC-motors for actuation, and a Raspberry Pi for computation. Duckietowns are structured modular urban environments with floor and signal layers. The floor layer includes five types of road segments: straight, curve, three-way intersection, four-way intersection, and empty. The signal layer includes traffic signs and traffic lights. Traffic lights have the same hardware as the Duckiebots, excluding the wheels, and are capable of sensing, computing, and actuation through their LEDs.

\subsubsection{The Duckietown software architecture} \label{sec:dt-sw}

We implement the Duckietown base software as ROS nodes and use the ROS topic messaging system for inter-process communication. The nodes are organized in Docker images that can be composed to satisfy various use-cases (Fig.~\ref{fig:dt-usecases}). Specifically, the components that correspond to the ``robot'' are the actuation and the sensing, and all other nodes correspond to the agent.
The Duckietown simulator~\cite{gym_duckietown} is a lightweight, modular, customizable virtual environment for rapid debugging and training of agents. It can replace the physical robot by subscribing to commands and producing sensor observations.
We provide an agent interface to the OpenAI Gym~\cite{brockman2016openai} protocol that can be called at a chosen frequency to speed up training.
In the case of reinforcement learning, rewards can be specified and are returned, enabling one to easily set up episodes for training.

The process of marshalling containers requires some specific configurations. To ease this process, we have developed the Duckietown Shell which is a wrapper around the Docker commands and provides an interface to the most frequent development and evaluation operations~\cite{dt-shell}. Additionally, the Duckietown Dashboard provides an intuitive high-level web-based graphical user interface~\cite{dt-dashboard}. The result is that agents that are developed in one modality (e.g., the simulator) can be evaluated in the cloud, on real robot hardware, or in the \ac{DTA} with one simple command or the at click of a button.

\subsection{System Architecture Overview}

The \ac{DUCKIENet} (Fig.~\ref{fig:duckienet-high-level}) is comprised of: (i) a Challenges server that stores agents, benchmarks, and results, computes leaderboards, and dispatches jobs to be executed by a distributed set of evaluators; (ii) a set of local or cloud-based evaluation machines that run the simulations; and (iii) one or more \acp{DTA}, physical instrumented labs that carry out physical experiments (Sec.~\ref{sec:dn-dta}).
The system has:
\begin{enumerate}
\item \textbf{Super-users} who define the benchmarks. The benchmarks are defined as Docker containers submitted to the Challenges server. These users can also set permissions for access to different benchmarks, their results, and the scores.
\item \textbf{Regular developers} are users of the system.
They develop agents locally and submit their agent code as Docker containers. They can observe the results of the evaluation, though the system supports notions of testing and validation, which hides some of the results.
\item \textbf{Simulation evaluators} and \textbf{experimental evaluators} query the Challenges server for jobs to execute. They run jobs as they receive them and communicate the results back to the Challenges server.
\end{enumerate}

\subsection{The Duckietown Automated Laboratory (\ac{DTA})} \label{sec:dn-dta}

    \begin{figure}[tb]
    \centering
        \includegraphics[trim = 0 0 0 -30, width=\columnwidth, clip]{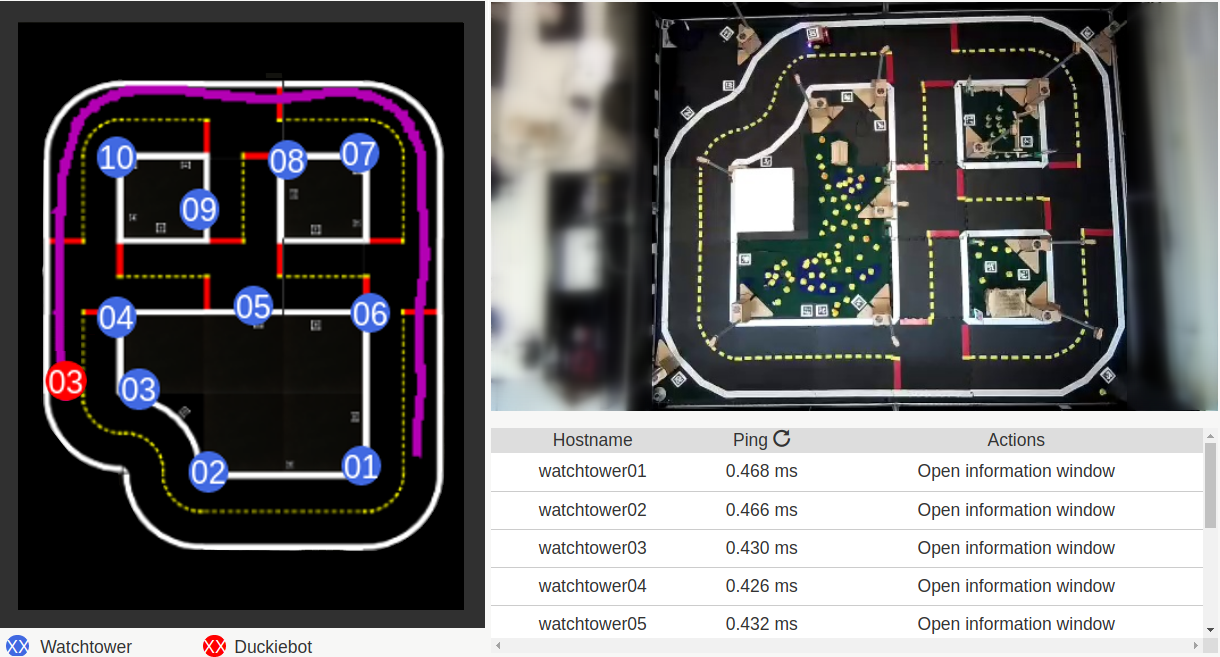}
        \caption{\textbf{The \ac{DTA} GUI} provides a control interface to human operators, including a representation of the map and the Watchtowers (left) and an overhead live feed of the physical Duckietown with diagnostic information about the experiment in progress (right). }
        \label{fig:dta-ui}

\end{figure}

The \acp{DTA} are remotely accessible hardware ecosystems designed to run reproducible experiments in a controlled environment with standardized hardware (Sec.~\ref{sec:dt-hw}). DTAs comprise: (a) a set of Duckietown environments (maps); (b) a localization system that estimates the pose of every Duckiebot in the map (Sec.~\ref{sec:dn-autolocalization}); (c) a local experiment manager, which receives jobs to execute from the Challenges server and coordinates the local Duckiebots, performs diagnostics, and collects logs from the system; (d) a human operator responsible for tasks that are currently not automated (e.g., resetting experiments and
``auto"-charging); and (e) a set of Duckiebots modified with a upwards-facing AprilTag~\cite{AprilTags} and charge collection apparatus, as shown in Figure~\ref{fig:db18-autolab}. A graphical UI (Fig.~\ref{fig:dta-ui}) makes the handling of experiments more user-friendly (Sec.~\ref{sec:dta-ui}).

We evaluate the reproducibility of the \acp{DTA} themselves by evaluating agents across different labs at distributed locations worldwide, as described in Section~\ref{sec:dn-validation}.

\begin{figure}[t]
\centering
    \includegraphics[trim = 0 0 0 -20,width=0.95\columnwidth]{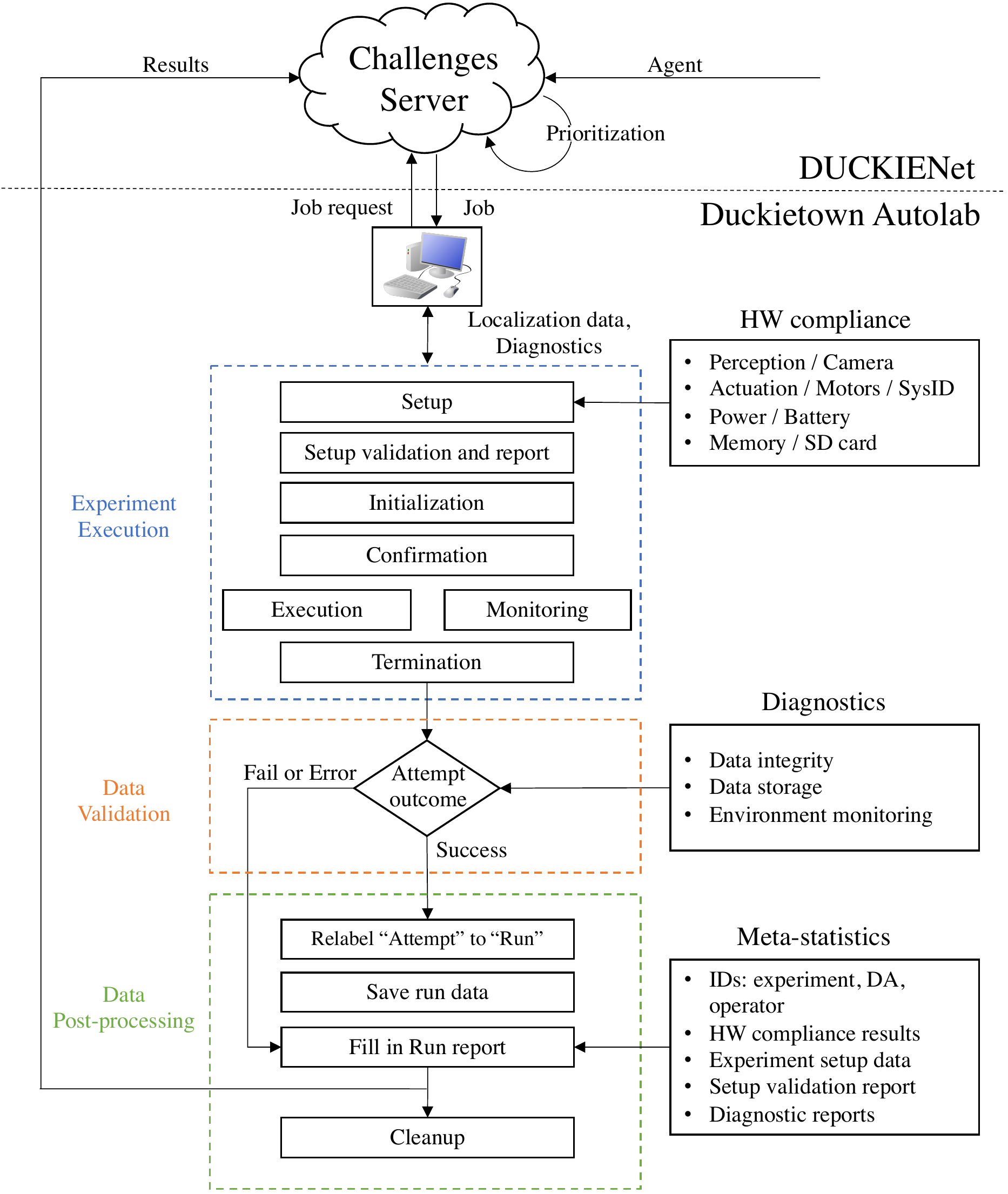}
    \caption{\textbf{Evaluation workflow}: An experiment is triggered by an agent being submitted to the Challenges server. The experiment is run in the \ac{DTA} following a predefined and deterministic procedure. The results are then uploaded to the Challenges server and returned to the submitter.}
    \label{fig:duckienet-hw-experiment}
\end{figure}

\subsubsection{Localization}
\label{sec:dn-autolocalization}

The goal of the localization system is to provide pose estimates of all Duckiebots during experiments. This data is post-processed according to metrics specified in the benchmarks to evaluate agent performance. The localization system comprises a network of Watchtowers, i.e., low-cost observation structures using the same sensing and computation as the Duckiebots.
The Watchtowers are placed such that their fields-of-view, which are restricted to a local neighbor region of each Watchtower, overlap and collectively cover the entire road network. A set of calibration AprilTags are fixed to the road layer of the map and provide reference points from the camera frame of each Watchtower to the map's fixed reference frame.

The localization system is decentralized and vision-based, using the camera both the streams of the network of Watchtowers and those of the Duckiebots themselves. In each image stream the system detects AprilTags, which are placed on the Duckiebots as well as in other locations around the city, e.g., on traffic signs at intersections. For each AprilTag, the system extracts their camera-relative pose. Using the resulting synchronized sequences of AprilTag-camera relative pose estimates, a pose graph-based estimate of the pose history of each AprilTag, using the $\text{g}^2 \text{o}$ library~\cite{g2o}, is returned.

\subsubsection{Operator console}
 \label{sec:dta-ui}

DTAs are equipped with a graphical user interface based on the \textbackslash compose\textbackslash~web framework (Fig.~\ref{fig:dta-ui}). Similarly to the Duckietown Dashboard, it is designed to provide an intuitive interface between the operator and the \ac{DTA}. Available functionalities include: various diagnostics (e.g., network latency, room illumination, camera feeds, etc.), the ability to run jobs (experiments) with a few clicks and compute, visualize and return results to the Challenges server, and the control of the environment illumination.

\subsection{Defining the Benchmarks}

Benchmarks can be defined either to be evaluated in simulation or on real robot hardware. Both types require the definition of:
(a) the environment, through a standardized map format;  (b) the \quotes{non-playing characters} and their behavior; and (c)  the evaluation metrics, which can be arbitrary functions defined on trajectories.

Each benchmark produces a set of scalar scores, where the designer can choose the final total order for scoring by describing a priority for the scores as well as tolerances (to define a lexicographic semiorder). The system allows one to define \quotes{benchmarks paths} as a directed acyclic graph of benchmarks with edges annotated with conditions for progressions. One simple use case is to prescribe that an agent be first evaluated in simulation before being evaluated on real robot hardware. It is possible to set thresholds on the various scores that need to be reached for progression. For competitions such as the AI Driving Olympics~\cite{aido}, the superusers can set up conditions such as \quotes{progress to the next stage only if the submission does better than a baseline}.

\subsection{DTA Operation Workflow}

The Challenges server sends a job to a \ac{DTA} that provides the instructions necessary to carry out the experiment (Fig.~\ref{fig:duckienet-hw-experiment}). The \ac{DTA} executes each experiment in a series of three steps: (i) experiment execution, (ii) data validation, and (iii) data post-processing and report generation.

Before the \ac{DTA} carries out an experiment, it first verifies that the robots pass a hardware compliance test that checks to see that their assembly, calibration, and the functionality of critical subsystems (sensing, actuation, power, and memory) are consistent with the specifications provided in the description of the experiment.

Having passed the compliance test, the experiment execution phase begins by initializing the pose of the robot(s) in the appropriate map, according to the experiment definition. The robot(s) that runs the user-defined agent is categorized as \quotes{active}, while the remaining \quotes{passive} robots run baseline agents. During the initialization step, the appropriate agents are transferred to each robot and initialized (as Docker containers). The localization system assigns a unique ID to each robot and verifies that the initial pose is correct. Once the robots are ready and the \ac{DTA} does not detect any abnormal conditions (e.g., related to room illumination and network performance), the \ac{DTA} sends a confirmation signal to synchronize the start of the experiment. The robots then execute their respective agents until a termination condition is met. Typical termination conditions include a robot exiting the road, crashing (i.e., not moving for longer than a threshold time), or reaching a time limit.

During the validation phase, the \ac{DTA} evaluates the integrity of the data, that it has been stored successfully, and that the experiment was not been terminated prematurely due to an invalidating condition (e.g., an unexpected change in environment lighting). Experiments that pass this phase are labeled as successful runs.

In the final phase, the \ac{DTA} compiles experiment data and passes it to the Challenges server. The Challenges server processes the data according to the specifications of the experiment and generates a report that summarizes the run, provides metrics and statistics unique to the run, as well as information that uniquely identifies the experiment, physical location, human operator, and hardware compliance. The entire report including the underlying data as well as optional comparative analysis with selected baselines are then shared with the user who submitted the agent. The Challenges server then performs cleanup in anticipation of the next job.

\subsection{The Artificial Intelligence Driving Olympics (AI-DO)}
\label{sec:applications-aido}

One of the key applications of our infrastructure is to host competitions.  The \acf{AI-DO}~\cite{aido} is a bi-annual competition that leverages the \ac{DUCKIENet} to compare the ability of various agents, ranging from traditional estimation-planning-control architectures to end-to-end learning-based architectures, to perform autonomous driving tasks of increasing complexity. By fixing the environment and the hardware, agents and tasks can be modified to benchmark performance across different challenges. Most robotics challenges offer \emph{either} embodied or simulated experiments. The \ac{DUCKIENet} enables seamless evaluation in \emph{both} along with a visualization of performance metrics, leaderboards, and access to the underlying raw data. The \ac{DUCKIENet} makes the process of participating accessible, with open-source baselines and detailed documentation.

\section{Validation} \label{sec:dn-validation}

We perform experiments to demonstrate (a) the repeatability of performance within a \ac{DTA}, (b) the reproducibility of experiments across different robots in the same \ac{DTA}, and (c) reproducibility across different facilities.

To assess the repeatability of an experiment, we run the same agent, in this case a \ac{LF} ROS-based baseline, on the same Duckiebot and in the same \ac{DTA}. To evaluate inter-robot reproducibility, we run the same experiment on multiple Duckiebots, keeping everything else fixed. Finally, we demonstrate that the same experiments yield similar results when run in different \ac{DTA}s across the world.

We define two metrics to measure repeatability: \ac{MPD} and \ac{MOD}. Given a set of trajectories, \ac{MPD} at a point along a trajectory is computed as the mean lateral displacement of the Duckiebot from the center of the lane. Similarly, \ac{MOD} represents the mean orientation with respect to the lane orientation. For each set of experiments, we compute the average and the standard deviation along the trajectory itself.

\subsection{Experiment Repeatability}

\begin{figure}[!tb]
        \expfigure{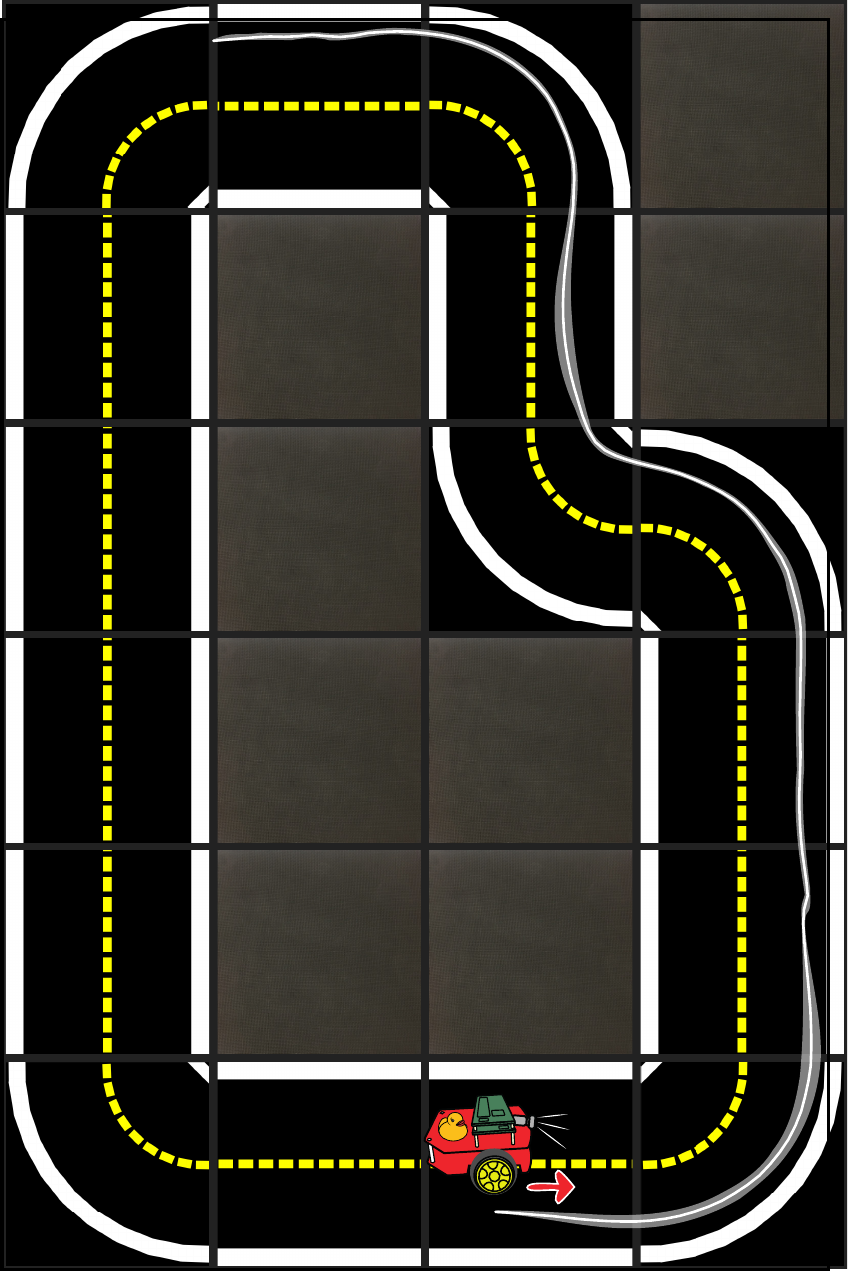}{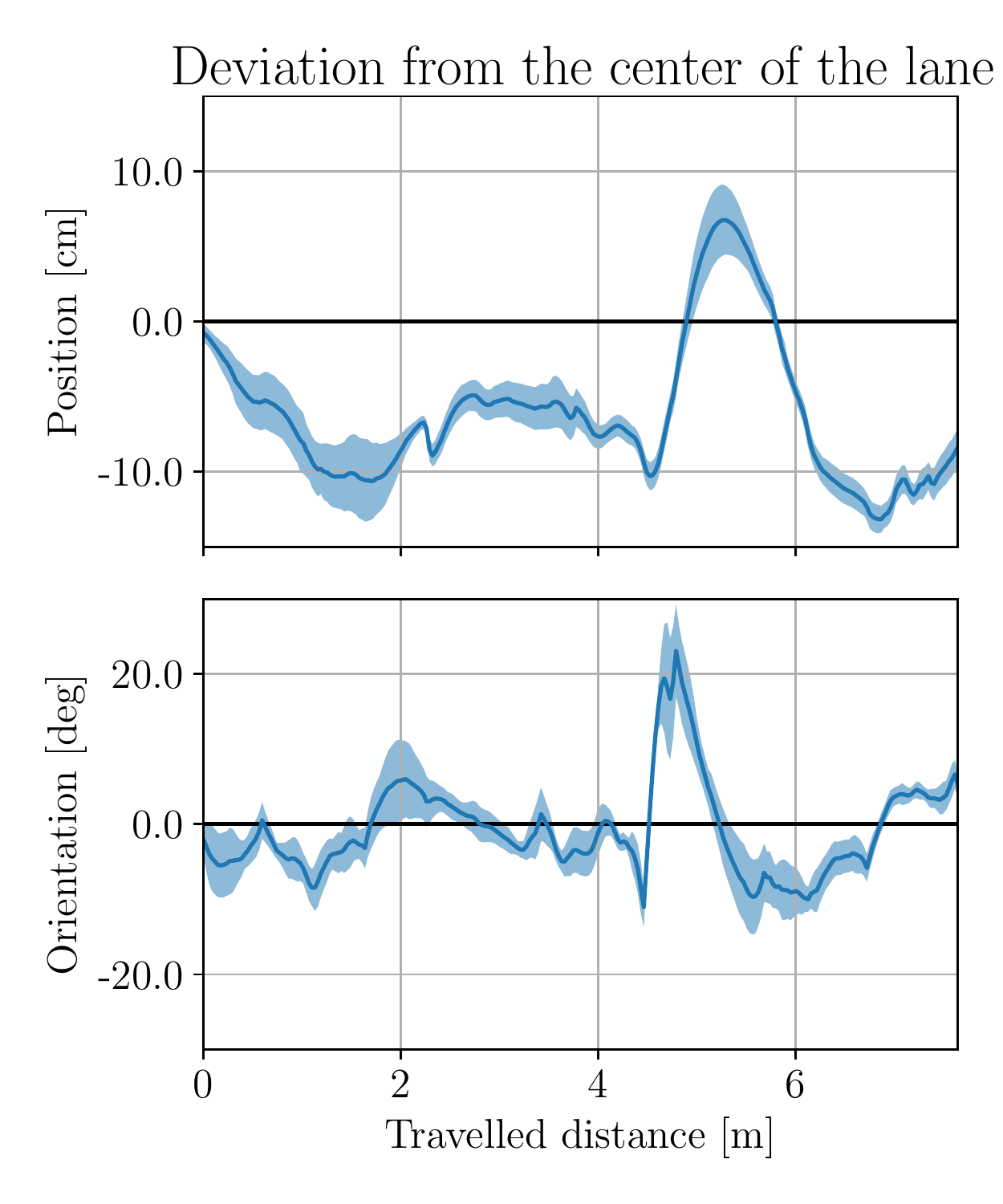}{\exptablesamebot}
        \caption{\textbf{Repeatability}: Experiments repeatedly run on the same robot. Plots include mean and standard deviation. The execution of the same agent is low variance across runs. }
        \label{fig:exp-reproducibility}
\end{figure}
To demonstrate experiment repeatability, we run a baseline agent for the \ac{LF} task nine times, on the same Duckiebot, with the same initial conditions and in the same \ac{DTA}. The results are in Figure~\ref{fig:exp-reproducibility}, with the mean and standard deviation of the robot's position on the map. We also include plots of the \ac{MPD} and \ac{MOD} metrics. Given the vetted robustness of the baseline agent, we expect repeatable behavior. This is supported by standard deviations of the \ac{MPD} ($1.3$\,cm) and the \ac{MOD} ($2.8$\,deg), which show that there is low variability in agent performance when run on the same hardware.

\subsection{Inter-Robot Reproducibility}

\begin{figure}[tb]
        \expfigure{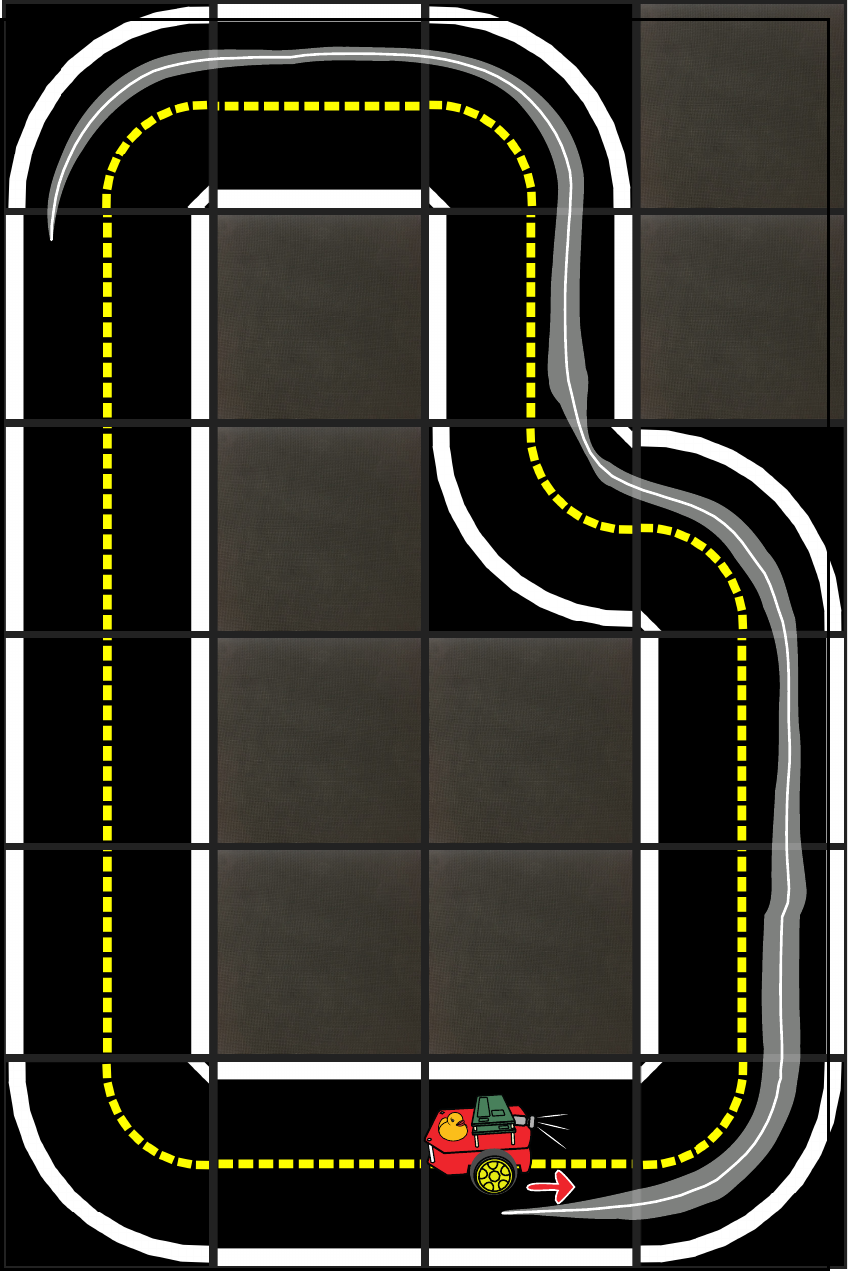}{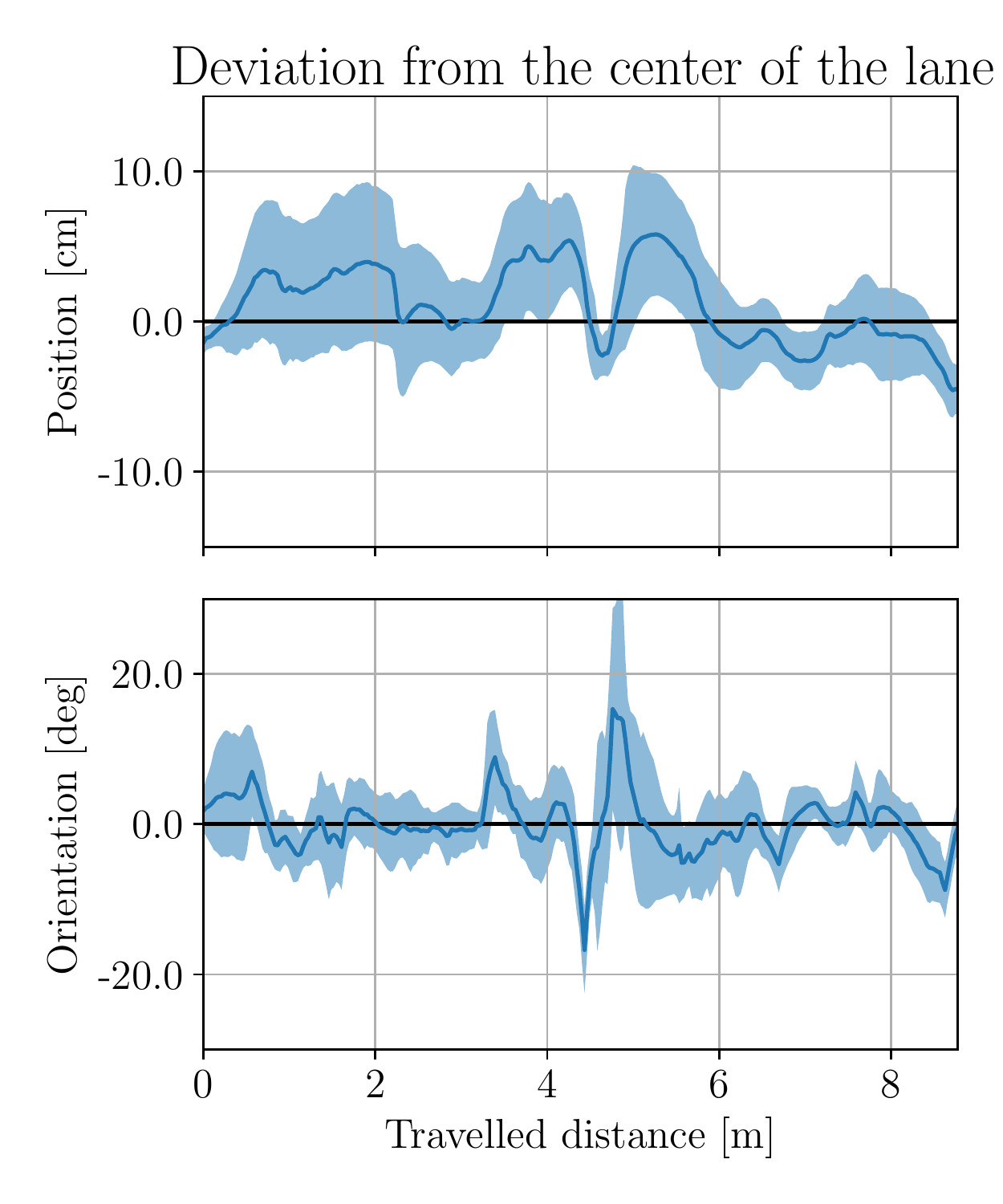}{\exptablediffbot}
        \caption{\textbf{Inter-Robot Reproducibility}: Experiments on three different but similar robots. Results show that there is more variation when the same agent is run on different hardware.}
        \label{fig:Duckiebot-reproducibility}
\end{figure}

Given the low-cost hardware setup, we expect a higher degree of variability if the same agent is run on different robots. To evaluate inter-robot reproducibility, we run the \ac{LF} baseline three times each, on three different Duckiebots. Experiments are nominally identical, and performed in the same \ac{DTA}. Although the behavior of each Duckiebot is expected to be repeatable given the result of the previous section, we expect some shift in the performance distribution to hardware nuisances such as slight variations in assembly, calibration, manufacturing differences of components, etc.

Figure~\ref{fig:Duckiebot-reproducibility} visualizes mean and standard deviation of the trajectories for all runs. The experiments reveal a standard deviation for the \ac{MPD} of $3.4$\,cm and a standard deviation for the \ac{MOD} of $5.2$\,deg. These results show higher deviations than for the single Duckiebot repeatability test, as expected.

\subsection{\ac{DTA} Reproducibility}

\begin{figure}
        \expfigure{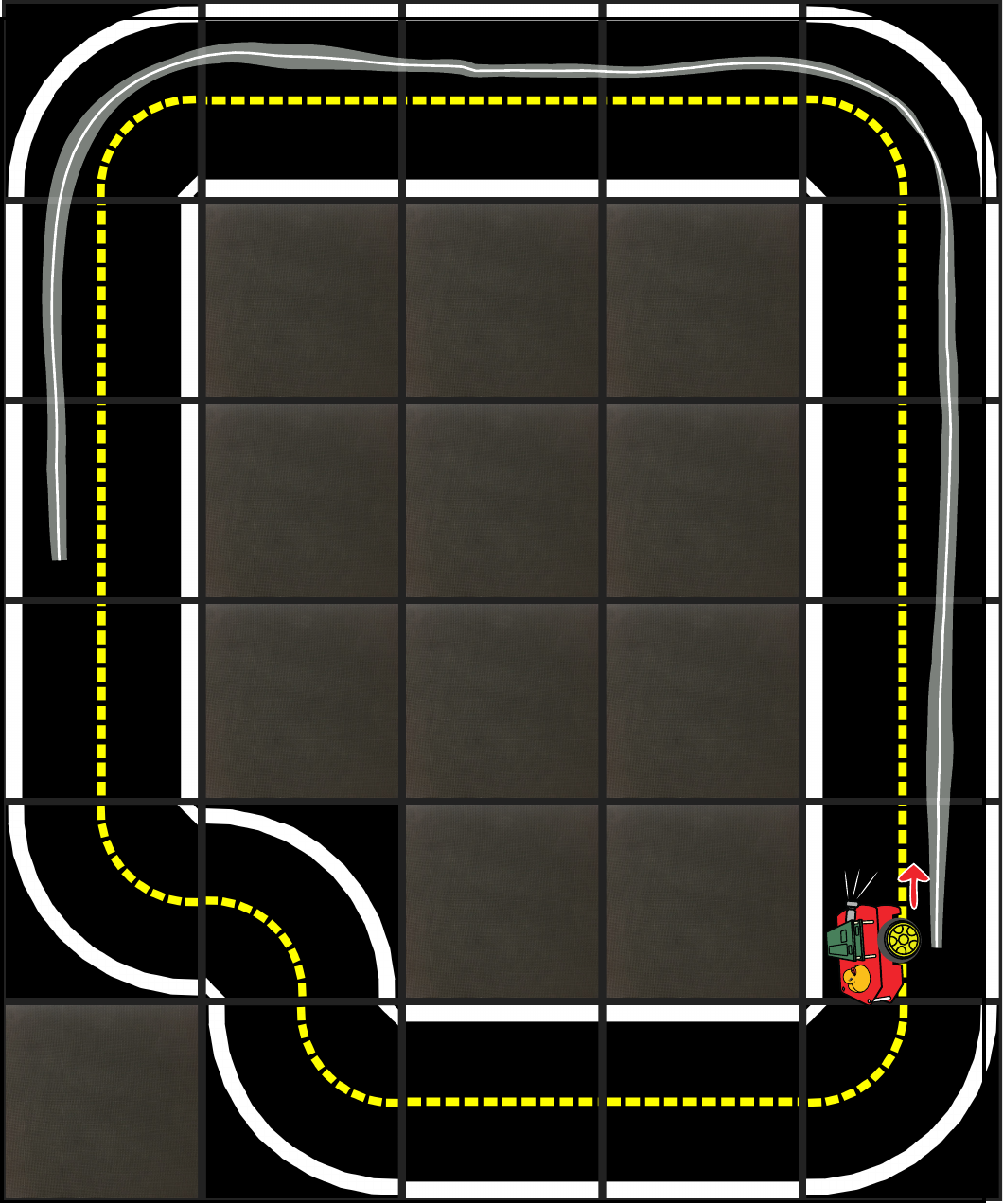}{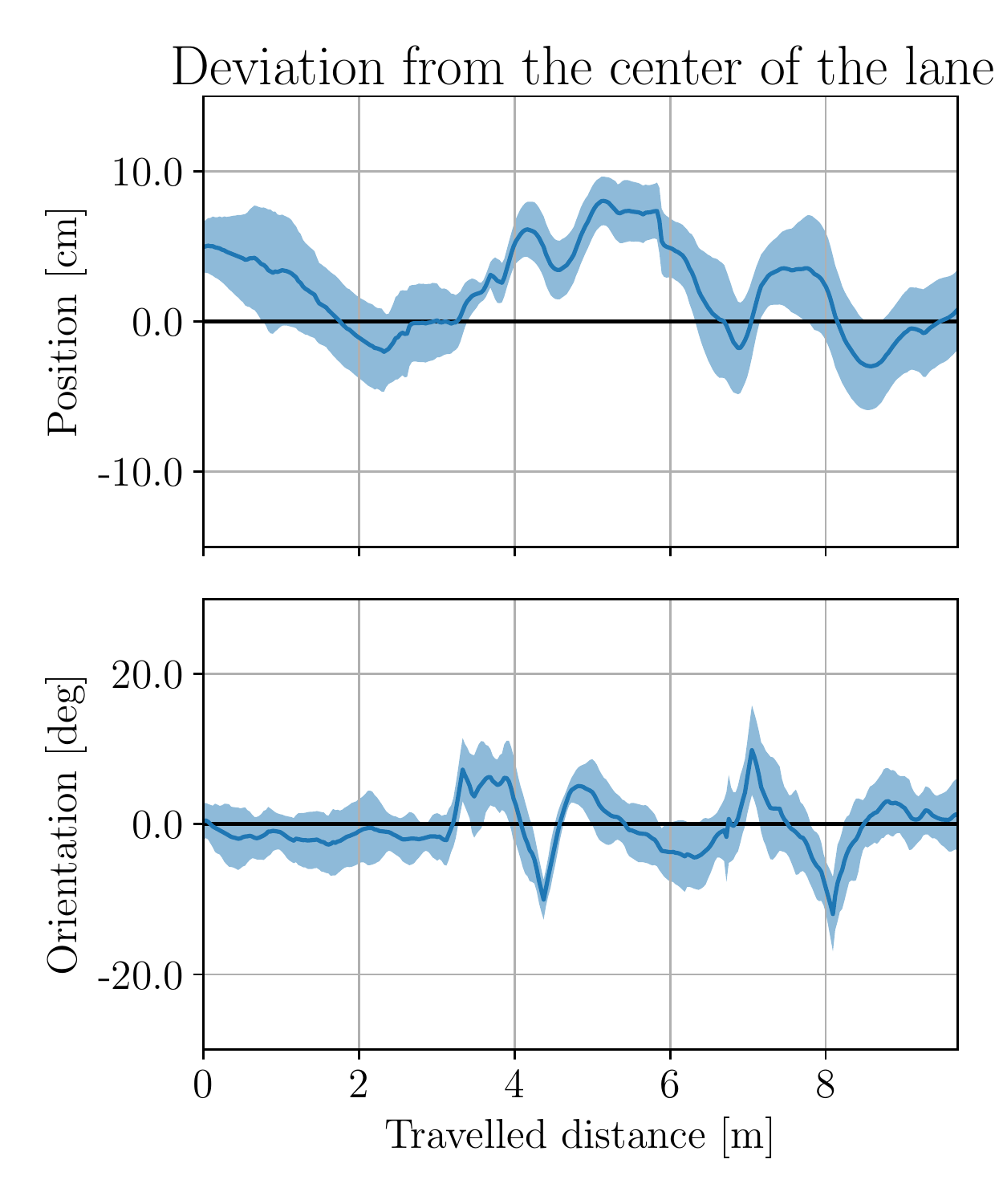}{\exptablediffsub}
        \caption{\textbf{DTA Reproducibility} Experiments in two different \acp{DTA}: ETH Z\"{u}rich and TTI-Chicago. Results show that the infrastructure is reproducible across setups, since experiments in two different \acp{DTA} yield similar results.}
        \label{fig:dta-reproducibility}
\end{figure}

To demonstrate \ac{DTA} reproducibility, we run a total of twelve experiments in two different \ac{DTA}s with nominally identical conditions except for the hardware, the geographic location, the infrastructure, and the human operators. Results are shown in Figure~\ref{fig:dta-reproducibility}. We obtain a standard deviation for the \ac{MPD} of $2.5$\,cm, and a standard deviation for the \ac{MOD} of $3.9$\,deg. Although variance is higher than the single Duckiebot repeatability test, which is to be expected, it is lower than that of the experiments run on three different robots, reinforcing the notion that hardware variability is measurable across different DTAs on the \ac{DUCKIENet}.

\subsection{Limitations}

Finally, we discuss some limitations of the \ac{DUCKIENet} framework, all of which are on our development roadmap.

The scenarios used to evaluate the reproducibility of the platform are relatively simple. With the Duckietown setup, we are able to produce much more complex behaviors and scenarios, such as multi-robot coordination, vehicle detection and avoidance, sign and signal detection, etc. These more complex behaviors should also be benchmarked. We have also only considered here metrics that are evaluated over a single agent, but multi-agent evaluation is also needed.

Finally, we have not analyzed the robustness to operator error (e.g., mis-configuration of the map compared to the simulation) or in case of hardware error (e.g., one camera in the localization system becomes faulty). This is necessary to encourage widespread adoption of the platform, which requires the components to be well-specified and capable of self-diagnosing configuration and hardware errors.

\section{Conclusions}
\label{sec:conclusion}

We have presented a framework for integrated robotics system design, development, and benchmarking. We subsequently presented a realization of this framework in the form the DUCKIENet, which comprises simulation, real robot hardware, flexible benchmark definitions, and remote evaluation. These components are swappable because they are designed to adhere to well-specified interfaces that define the abstractions.
In order to achieve this level of reproducibility, we have relied heavily on concepts and technologies including formal specifications and software containerization.
The combination of these tools with the proper abstractions and interfaces yields a procedure that can be followed by other roboticists in a straightforward manner.

Our contention is that there is a need for stronger efforts towards reproducible research for robotics, and that to achieve this we need to consider the evaluation in equal terms as the algorithms themselves. In this fashion, we can obtain reproducibility \emph{by design} through the research and development processes. Achieving this on a large-scale will contribute to a more systemic evaluation of robotics research and, in turn, increase the progress of development.

\section*{Acknowledgements}

Liam Paull and Anthony Courchesne are supported by CIFAR (CCAI Chair Program), NSERC (Discovery Grant Program).  Bhairav Mehta is additionally supported by Mila and IVADO Excellence Scholarship Fund. Andrea F.\ Daniele and Matthew R.\ Walter are supported in part by the National Science Foundation under grant IIS-1638072 and by the Robotics Consortium of the U.S. Army Research Laboratory under the Collaborative Technology Alliance Program Cooperative Agreement W911NF-10-2-0016.

We also thank Josefine Quack, Eric Lu, Ben Weber, Florian Golemo, Chude Qian, all students who ran experiments on the system, as well as Marc-Andre Corzillius, Marcus Aaltonen, and everyone at ETH's Institute for Dynamic Systems and Control (IDSC) who supported us.

\bibliographystyle{IEEEtran}

\bibliography{99_references}

\begin{thebibliography}{10}
\providecommand{\url}[1]{#1}
\csname url@samestyle\endcsname
\providecommand{\newblock}{\relax}
\providecommand{\bibinfo}[2]{#2}
\providecommand{\BIBentrySTDinterwordspacing}{\spaceskip=0pt\relax}
\providecommand{\BIBentryALTinterwordstretchfactor}{4}
\providecommand{\BIBentryALTinterwordspacing}{\spaceskip=\fontdimen2\font plus
\BIBentryALTinterwordstretchfactor\fontdimen3\font minus
  \fontdimen4\font\relax}
\providecommand{\BIBforeignlanguage}[2]{{%
\expandafter\ifx\csname l@#1\endcsname\relax
\typeout{** WARNING: IEEEtran.bst: No hyphenation pattern has been}%
\typeout{** loaded for the language `#1'. Using the pattern for}%
\typeout{** the default language instead.}%
\else
\language=\csname l@#1\endcsname
\fi
#2}}
\providecommand{\BIBdecl}{\relax}
\BIBdecl

\bibitem{paull2017duckietown}
L.~Paull, J.~Tani, H.~Ahn, J.~Alonso-Mora, L.~Carlone, M.~Cap, Y.~F. Chen,
  C.~Choi, J.~Dusek, Y.~Fang \emph{et~al.}, ``Duckietown: an open, inexpensive
  and flexible platform for autonomy education and research,'' in
  \emph{Proceedings of the IEEE International Conference on Robotics and
  Automation (ICRA)}, 2017, pp. 1497--1504.

\bibitem{Henderson2017}
P.~Henderson, R.~Islam, P.~Bachman, J.~Pineau, D.~Precup, and D.~Meger, ``Deep
  reinforcement learning that matters,'' in \emph{Proceedings of the National
  Conference on Artificial Intelligence (AAAI)}, 2018, pp. 3207--3214.

\bibitem{kitti}
A.~Geiger, P.~Lenz, and R.~Urtasun, ``{Are we ready for autonomous driving?
  {The KITTI} vision benchmark suite},'' in \emph{Proceedings of the IEEE
  Conference on Computer Vision and Pattern Recognition (CVPR)}, 2012, pp.
  3354--3361.

\bibitem{cityscapes}
M.~Cordts, M.~Omran, S.~Ramos, T.~Rehfeld, M.~Enzweiler, R.~Benenson,
  U.~Franke, S.~Roth, and B.~Schiele, ``The {C}ityscapes dataset for semantic
  urban scene understanding,'' in \emph{Proceedings of the IEEE Conference on
  Computer Vision and Pattern Recognition (CVPR)}, 2016, pp. 3213--3223.

\bibitem{nuscenes}
H.~Caesar, V.~Bankiti, A.~H. Lang, S.~Vora, V.~E. Liong, Q.~Xu, A.~Krishnan,
  Y.~Pan, G.~Baldan, and O.~Beijbom, ``{NuScenes}: {A} multimodal dataset for
  autonomous driving,'' \emph{arXiv preprint arXiv:1903.11027}, 2019.

\bibitem{waymo-sun2019scalability}
P.~Sun, H.~Kretzschmar, X.~Dotiwalla, A.~Chouard, V.~Patnaik, P.~Tsui, J.~Guo,
  Y.~Zhou, Y.~Chai, B.~Caine \emph{et~al.}, ``Scalability in perception for
  autonomous driving: {W}aymo open dataset,'' in \emph{Proceedings of the IEEE
  Conference on Computer Vision and Pattern Recognition (CVPR)}, 2020, pp.
  2446--2454.

\bibitem{yu2020introduction}
Y.~Yu, Z.~Liu, and X.~Wang, ``An introduction to formation control of {UAV}
  with {V}icon system,'' in \emph{Proceedings of the International Conference
  on Robotic Sensor Networks (ROSENET)}, 2020, pp. 181--190.

\bibitem{robotarium}
D.~Pickem, P.~Glotfelter, L.~Wang, M.~Mote, A.~Ames, E.~Feron, and
  M.~Egerstedt, ``{The {R}obotarium: A remotely accessible swarm robotics
  research testbed},'' in \emph{Proceedings of the IEEE International
  Conference on Robotics and Automation (ICRA)}, 2017, pp. 1699--1706.

\bibitem{vicon-merriaux2017study}
P.~Merriaux, Y.~Dupuis, R.~Boutteau, P.~Vasseur, and X.~Savatier, ``A study of
  {V}icon system positioning performance,'' \emph{Sensors}, vol.~17, no.~7,
  2017.

\bibitem{dosovitskiy2017carla}
A.~Dosovitskiy, G.~Ros, F.~Codevilla, A.~Lopez, and V.~Koltun, ``{CARLA}: {A}n
  open urban driving simulator,'' \emph{arXiv preprint arXiv:1711.03938}, 2017.

\bibitem{shah2018airsim}
S.~Shah, D.~Dey, C.~Lovett, and A.~Kapoor, ``{AirSim}: High-fidelity visual and
  physical simulation for autonomous vehicles,'' in \emph{Proceedings of the
  International Conference on Field and Service Robotics (FSR)}, 2018, pp.
  621--635.

\bibitem{krishnan2019air}
S.~Krishnan, B.~Borojerdian, W.~Fu, A.~Faust, and V.~J. Reddi, ``Air learning:
  An {AI} research platform for algorithm-hardware benchmarking of autonomous
  aerial robots,'' \emph{arXiv preprint arXiv:1906.00421}, 2019.

\bibitem{amigoni2015competitions}
F.~Amigoni, E.~Bastianelli, J.~Berghofer, A.~Bonarini, G.~Fontana,
  N.~Hochgeschwender, L.~Iocchi, G.~Kraetzschmar, P.~Lima, M.~Matteucci
  \emph{et~al.}, ``Competitions for benchmarking: Task and functionality
  scoring complete performance assessment,'' \emph{IEEE Robotics \& Automation
  Magazine}, vol.~22, no.~3, pp. 53--61, 2015.

\bibitem{buehler2009darpa}
M.~Buehler, K.~Iagnemma, and S.~Singh, \emph{The {DARPA} urban challenge:
  {A}utonomous vehicles in city traffic}.\hskip 1em plus 0.5em minus
  0.4em\relax Springer, 2009, vol.~56.

\bibitem{johnson2015team}
M.~Johnson, B.~Shrewsbury, S.~Bertrand, T.~Wu, D.~Duran, M.~Floyd, P.~Abeles,
  D.~Stephen, N.~Mertins, A.~Lesman \emph{et~al.}, ``Team {IHMC}'s lessons
  learned from the {DARPA} robotics challenge trials,'' \emph{Journal of Field
  Robotics}, vol.~32, no.~2, pp. 192--208, 2015.

\bibitem{correll2016analysis}
N.~Correll, K.~E. Bekris, D.~Berenson, O.~Brock, A.~Causo, K.~Hauser, K.~Okada,
  A.~Rodriguez, J.~M. Romano, and P.~R. Wurman, ``Analysis and observations
  from the first {A}mazon {P}icking {C}hallenge,'' \emph{IEEE Transactions on
  Automation Science and Engineering}, vol.~15, no.~1, pp. 172--188, 2016.

\bibitem{pickem2017robotarium}
D.~Pickem, P.~Glotfelter, L.~Wang, M.~Mote, A.~Ames, E.~Feron, and
  M.~Egerstedt, ``The {R}obotarium: {A} remotely accessible swarm robotics
  research testbed,'' in \emph{Proceedings of the IEEE International Conference
  on Robotics and Automation (ICRA)}, 2017, pp. 1699--1706.

\bibitem{wilson2020robotarium}
S.~Wilson, P.~Glotfelter, L.~Wang, S.~Mayya, G.~Notomista, M.~Mote, and
  M.~Egerstedt, ``The {R}obotarium: Globally impactful opportunities,
  challenges, and lessons learned in remote-access, distributed control of
  multirobot systems,'' \emph{IEEE Control Systems Magazine}, vol.~40, no.~1,
  pp. 26--44, 2020.

\bibitem{Goodman341ps12}
\BIBentryALTinterwordspacing
S.~N. Goodman, D.~Fanelli, and J.~P.~A. Ioannidis, ``What does research
  reproducibility mean?'' \emph{Science Translational Medicine}, vol.~8, no.
  341, 2016. [Online]. Available:
  \url{https://stm.sciencemag.org/content/8/341/341ps12}
\BIBentrySTDinterwordspacing

\bibitem{blischak2016quick}
J.~D. Blischak, E.~R. Davenport, and G.~Wilson, ``A quick introduction to
  version control with {G}it and {G}it{H}ub,'' \emph{PLoS Computational
  Biology}, vol.~12, no.~1, 2016.

\bibitem{boettiger2015introduction}
C.~Boettiger, ``An introduction to {D}ocker for reproducible research,''
  \emph{ACM SIGOPS Operating Systems Review}, vol.~49, no.~1, pp. 71--79, 2015.

\bibitem{Docker}
\BIBentryALTinterwordspacing
I.~Docker. Docker: Empowering app development for developers. [Online].
  Available: \url{https://www.docker.com}
\BIBentrySTDinterwordspacing

\bibitem{weisz2016robobench}
J.~Weisz, Y.~Huang, F.~Lier, S.~Sethumadhavan, and P.~Allen, ``Robobench:
  Towards sustainable robotics system benchmarking,'' in \emph{Proceedings of
  the IEEE International Conference on Robotics and Automation (ICRA)}, 2016,
  pp. 3383--3389.

\bibitem{ROS}
M.~Quigley, K.~Conley, B.~P. Gerkey, J.~Faust, T.~Foote, J.~Leibs, R.~Wheeler,
  and A.~Y. Ng, ``{ROS}: {A}n open-source {R}obot {O}perating {S}ystem,'' in
  \emph{International Concerence on Robotics and Automation (ICRA) Workshop on
  Open Source Software}, 2009.

\bibitem{huang10}
A.~S. Huang, E.~Olson, and D.~Moore, ``{LCM}: Lightweight communications and
  marshalling,'' in \emph{Proceedings of the IEEE/RSJ International Conference
  on Intelligent Robots and Systems (IROS)}, Taipei, Taiwan, October 2010, pp.
  4057--4062.

\bibitem{cbor}
``{CBOR},'' \url{http://cbor.io/}, accessed: 2020-02-29.

\bibitem{tani2016duckietown}
J.~Tani, L.~Paull, M.~T. Zuber, D.~Rus, J.~How, J.~Leonard, and A.~Censi,
  ``Duckietown: {A}n innovative way to teach autonomy,'' in \emph{International
  Conference EduRobotics 2016}, 2016, pp. 104--121.

\bibitem{aido}
J.~Zilly, J.~Tani, B.~Considine, B.~Mehta, A.~F. Daniele, M.~Diaz,
  G.~Bernasconi, C.~Ruch, J.~Hakenberg, F.~Golemo \emph{et~al.}, ``The {AI}
  {D}riving {O}lympics at {N}eur{IPS} 2018,'' in \emph{The NeurIPS'18
  Competition}.\hskip 1em plus 0.5em minus 0.4em\relax Springer, 2020, pp.
  37--68.

\bibitem{dt-org}
``{D}uckietown project,'' \url{https://www.duckietown.org/}, accessed:
  2020-01-22.

\bibitem{gym_duckietown}
M.~Chevalier-Boisvert, F.~Golemo, Y.~Cao, B.~Mehta, and L.~Paull, ``Duckietown
  environments for {O}pen{AI} {G}ym,''
  \url{https://github.com/duckietown/gym-duckietown}, 2018.

\bibitem{brockman2016openai}
G.~Brockman, V.~Cheung, L.~Pettersson, J.~Schneider, J.~Schulman, J.~Tang, and
  W.~Zaremba, ``Open{AI} {G}ym,'' \emph{arXiv preprint arXiv:1606.01540}, 2016.

\bibitem{dt-shell}
``The {D}uckietown shell,''
  \url{https://github.com/duckietown/duckietown-shell}, accessed: 2020-03-01.

\bibitem{dt-dashboard}
``The {D}uckietown dashboard,''
  \url{https://docs.duckietown.org/DT19/opmanual_duckiebot/out/duckiebot_dashboard_setup.html},
  accessed: 2020-03-01.

\bibitem{AprilTags}
E.~Olson, ``{AprilTag: A robust and flexible visual fiducial system},'' in
  \emph{IEEE International Conference on Robotics and Automation (ICRA)},
  Shanghai, China, 2011, pp. 3400--3407.

\bibitem{g2o}
G.~Grisetti, R.~K{\"u}mmerle, H.~Strasdat, and K.~Konolige,
  ``$\text{g}^2\text{o}$: {A} general framework for (hyper) graph
  optimization,'' in \emph{IEEE International Conference on Robotics and
  Automation (ICRA), Shanghai, China}, 2011, pp. 9--13.

\end{thebibliography}

\end{document}